\documentclass[pdflatex,sn-mathphys-ay,hidelinks]{sn-jnl}

\usepackage[utf8]{inputenc}
\usepackage{graphicx}
\usepackage{mathrsfs}
\usepackage[title]{appendix}
\usepackage{xcolor}
\usepackage{textcomp}
\usepackage{manyfoot}
\usepackage{algorithm}
\usepackage{algorithmic}
\usepackage{booktabs}
\usepackage{amsfonts}
\usepackage{nicefrac}
\usepackage{microtype}
\usepackage{subcaption}
\usepackage{siunitx}
\usepackage{multirow}
\usepackage{amsmath}
\usepackage{amssymb}
\usepackage{mathtools}
\usepackage{amsthm}
\usepackage{dsfont}
\usepackage[hyphenbreaks]{breakurl}

\mathtoolsset{showonlyrefs}
\theoremstyle{plain}
\newtheorem{theorem}{Theorem}[section]
\newtheorem{proposition}[theorem]{Proposition}

\theoremstyle{definition}

\theoremstyle{remark}

\newtheorem{example}[theorem]{Example}

\DeclareMathOperator{\X}{\mathcal{X}}
\DeclareMathOperator{\Y}{\mathcal{Y}}
\DeclareMathOperator{\D}{\mathcal{D}}

\DeclareMathOperator{\kl}{KL}
\DeclareMathOperator{\dirich}{Dir}
\DeclareMathOperator{\lagrange}{L}
\DeclareMathOperator{\loss}{\mathcal{L}}
\DeclareMathOperator{\risk}{\mathcal{R}}

\DeclareMathOperator{\expect}{\mathbb{E}}
\DeclareMathOperator{\var}{Var}
\DeclareMathOperator{\sm}{softmax}
\DeclareMathOperator{\R}{\mathbb{R}}
\DeclareMathOperator{\N}{\mathbb{N}}

\DeclareMathOperator{\sgn}{sign}
\DeclareMathOperator{\lbl}{\ell}

\newcommand\vc[1]{\boldsymbol{#1}}
\newcommand\sublog[1]{\mathfrak{#1}}

\begin{document}

\title{Robust Partial-Label Learning by Leveraging Class Activation Values}

\author{\fnm{Tobias} \sur{Fuchs}}\email{tobias.fuchs@kit.edu}

\author{\fnm{Florian} \sur{Kalinke}}\email{florian.kalinke@kit.edu}

\affil{\orgname{Karlsruhe Institute of Technology}, \orgaddress{\country{Germany}}}


\abstract{
    Real-world training data is often noisy; for example, human annotators assign conflicting class labels to the same instances.
    Partial-label learning (PLL) is a weakly supervised learning paradigm that allows training classifiers in this context without manual data cleaning.
    While state-of-the-art methods have good predictive performance, their predictions are sensitive to high noise levels, out-of-distribution data, and adversarial perturbations.
    We propose a novel PLL method based on subjective logic, which explicitly represents uncertainty by leveraging the magnitudes of the underlying neural network's class activation values.
    Thereby, we effectively incorporate prior knowledge about the class labels by using a novel label weight re-distribution strategy that we prove to be optimal.
    We empirically show that our method yields more robust predictions in terms of predictive performance under high PLL noise levels, handling out-of-distribution examples, and handling adversarial perturbations on the test instances.
}

\keywords{Weakly-Supervised Learning, Partial-Label Learning, Learning under Noise, Robust Classification, Deep Learning}



\maketitle

\section{Introduction}
\label{sec:intro}
In real-world applications, one often encounters ambiguously labeled data.
In crowd labeling, for example, human annotation produces instances with multiple conflicting class labels \citep{HuiskesL08}.
Other examples with ambiguous data include web mining \citep{GuillauminVS10,ZengXJCGXM13} and audio classification \citep{BriggsFR12}.
Partial-label learning (PLL; \citealt{grandvalet2002logistic,NguyenC08,ZhangYT17,0009LLQG23}) is a weakly-supervised learning paradigm that targets classification with such inexact supervision, where training data can have several candidate labels of which only one is correct.
While cleaning is costly, PLL algorithms allow to handle such ambiguously labeled data directly.

As predictions by machine-learning systems often impact actions or decisions by humans, they should be \emph{robust} regarding several criteria to limit mispredictions and their effects.
Three common criteria are robustness against (a) high noise levels \citep{ZhangZW0021}, (b) out-of-distribution data \citep{SensoyKK18}, and (c) adversarial attacks \citep{MadryMSTV18}.
Consider, for example, safety-critical domains such as medical image classification \citep{YangWMLC09,LambrouPG11,ReamaroonSLIN19} or financial fraud detection~\citep{ChengXSZY020,BerkmansK23,XiangZCLZOC023}, where all three criteria are of interest.

Robustness in terms of (a), (b), and (c) is especially important in PLL because of its noisy and inexact supervision.
While different noise generation processes (a) are well-examined in PLL, it is still open to investigate the impact of (b) and (c) on PLL algorithms.
Dealing with out-of-distribution data (b) is essential in the web mining use case of PLL as the closed-world assumption usually does not hold, that is, an algorithm should recognize instances that do not belong to any known class.
Addressing adversarial modifications of input features (c) is also critical, given that much of the training data is human-based, presenting a potential vulnerability.
Tackling (a)\,--\,(c) is particularly challenging in the PLL domain as there is no exact ground truth on which an algorithm can rely to build robust representations.
Our proposed PLL method is unique in its ability to perform well across all three aspects.

In this work, we propose a novel PLL deep-learning algorithm that leverages the magnitudes of class activation values within the subjective logic framework \citep{Josang16}.
Subjective logic allows for explicitly representing uncertainty in predictions, which is highly beneficial in dealing with the challenges (a)\,--\,(c).
The details are as follows.
When dealing with noise from the PLL candidate sets (a), having good uncertainty estimates supports the propagation of meaningful labeling information as it allows us to put more weight on class labels with a low uncertainty and to restrict the influence of noisy class labels, which have high uncertainty.
We can tackle out-of-distribution data (b) by optimizing for high uncertainty when the correct class label is excluded from the set of all possible class labels.
Adversarial modifications of the input features (c) are addressed similarly to (a), as our approach provides reliable uncertainty estimates near the decision boundaries of the class labels.

The supervised classification approach by \citet{SensoyKK18} is the most similar to the proposed approach as both employ the subjective logic framework.
However, it is highly non-trivial to extend the methods from the supervised to the PLL setting as the existing work relies on exact ground truth, which is generally unavailable in PLL.
We attack this problem by proposing a novel representation of partially-labeled data within the subjective logic framework and give an optimal update strategy for the candidate label weights with respect to the model's loss term.
Subjective logic allows us to deal with the partially labeled data in a principled fashion by jointly learning the candidate labels' weights and their associated uncertainties.

Our \textbf{contributions} are as follows.
\begin{itemize}
    \item We introduce \textsc{RobustPll}, a novel partial-label learning algorithm, which leverages the model's class activation values within the subjective logic framework.
    \item We empirically demonstrate that \textsc{RobustPll} yields more robust predictions than our competitors.
          The proposed method achieves state-of-the-art prediction performance under high PLL noise and can deal with out-of-distribution examples and examples corrupted by adversarial noise more reliably.
          Our code and data are publicly available.\footnote{\label{fnt:code}\url{https://github.com/mathefuchs/robust-pll}}
    \item Our analysis of \textsc{RobustPll} shows that the proposed label weight update strategy is optimal in terms of the mean-squared error and allows for reinterpretation within the subjective logic framework.
          Further, we discuss our method's runtime and show that it yields the same runtime complexity as other state-of-the-art PLL algorithms.
\end{itemize}

\textbf{Structure of the paper.}
Section~\ref{sec:related-work} discusses related work. Section~\ref{sec:prelim-pll} defines the problem and our notations.
We propose our PLL method in Section~\ref{sec:main} and show our experiments in Section~\ref{sec:exp}. Section~\ref{sec:conclusions} concludes.
We defer all proofs and hyperparameter choices to the appendices.

\section{Related Work}
\label{sec:related-work}
This section separately details related work on PLL and on making predictions more robust regarding aspects (a)\,--\,(c).

\subsection{Partial-Label Learning (PLL)}
PLL is a typical weakly-supervised learning problem.
Early approaches apply common supervised learning frameworks to the PLL context:
\citet{grandvalet2002logistic} propose a logistic regression formulation,
\citet{JinG02} propose an expectation-maximization strategy,
\citet{HullermeierB06} propose a nearest-neighbors method,
\citet{NguyenC08} propose an extension of support-vector machines,
and \citet{CourST11} introduce an average loss formulation allowing the use of any supervised method.

As most of the aforementioned algorithms struggle with non-uniform noise, several extensions and novel methods have been proposed:
\citet{ZhangY15a,ZhangZL16,XuLG19,WangLZ19,FengA19,NiZDCL21} leverage ideas from representation learning,
\citet{YuZ17,WangLZ19,FengA19,NiZDCL21} extend the maximum-margin idea,
\citet{LiuD12,LvXF0GS20} propose extensions of the expectation-maximization strategy,
\citet{ZhangYT17,TangZ17,WuZ18} propose stacking and boosting ensembles,
and \citet{LvXF0GS20,CabannesRB20} introduce a minimum loss formulation, which iteratively disambiguates the partial labels.

The progress of deep-learning techniques also yields advances in PLL.
\citet{FengL0X0G0S20,LvXF0GS20} provide risk-consistent loss formulations for PLL and \citet{XuQGZ21,WangXLF0CZ22,HeFLLY22,ZhangF0L0QS22,0009LLQG23,fan2024kmt,crosel2024} use advances in deep representation learning and data augmentation to strengthen inference from PLL data.
Similar to our approach, \textsc{Cavl} \citep{ZhangF0L0QS22} makes use of the class activation values to identify the correct labels in the candidate sets.
While they use the activation values as a feature map, we use the activation values to build multinomial opinions in subjective logic, which reflect the involved uncertainty in prediction-making.

Similar to \textsc{Proden} \citep{LvXF0GS20}, \textsc{Pop} \citep{0009LLQG23}, and \textsc{CroSel} \citep{crosel2024}, among many others, we iteratively update a label weight vector keeping track of the model's knowledge about the labeling of all instances.
However, those three methods do not provide any formal reasoning for their respective update rules.
In contrast, we prove our update rule's optimality in Proposition~\ref{prop:optimal} and Proposition~\ref{prop:optimal-ce} for the mean-squared error and cross-entropy error, respectively.

We note that, at a first glance, our update rule also appears to be similar to the label smoothing proposed by \citet{GongB024}.
Based on a smoothing hyperparameter~$r \in [0, 1]$, they iteratively update the label weights: $r = 1$ uniformly allocates weight among all class labels, while $r = 0$ only allocates label weight on the most likely label.
In contrast, our update strategy does not involve any hyperparameter and allocates probability mass based on the uncertainty involved in prediction-making.

\subsection{Robust Prediction-Making}
Robust prediction-making encompasses a variety of aspects out of which we consider (a) good predictive performance under high PLL noise \citep{ZhangZW0021}, (b) robustness against out-of-distribution examples (OOD; \citealt{SensoyKK18}), and (c) robustness against adversarial examples \citep{MadryMSTV18} to be the most important in PLL.
Real-world applications of PLL often entail web mining use cases, where the closed-world assumption usually does not hold (requiring (b)).
Also, PLL training data is commonly human-based and therefore a possible surface for adversarial attacks (requiring (c)).
Other robustness objectives that we do \emph{not} consider are, for example, the decomposition of the involved uncertainties \citep{KendallG17,WimmerSHBH23} or the calibration of the prediction confidences \citep{AoRS23,MortierBHLW23}.

To address (a) in the supervised setting, one commonly employs Bayesian methods \citep{KingmaW13,KendallG17} or ensembles\footnote{Ensemble techniques also benefit (b) and (c) and are easy to implement. Therefore, we also consider an ensemble approach of one of our competitors in our experiments as a strong baseline in Section~\ref{sec:exp}.} \citep{Lakshminarayanan17,WimmerSHBH23}.
To recognize OOD samples (b), one commonly employs techniques from representation learning \citep{ZhangY15a,XuQGZ21} or leverages negative examples using regularization or contrastive learning \citep{SensoyKK18,WangXLF0CZ22}.
To address (c), methods incorporate adversarially corrupted features already in the training process to strengthen predictions \citep{Lakshminarayanan17}.
To the best of our knowledge, we are the first to propose a method that addresses (a), (b), and (c) in PLL.
Tackling all three aspects is particularly challenging in the PLL domain as there is no exact ground truth on which an algorithm can rely to build robust representations.

\section{Problem Statement and Notations}
\label{sec:prelim-pll}
This section defines the partial-label learning problem, establishes the notations used throughout this work, and briefly summarizes subjective logic.

\subsection{Partial-Label Learning (PLL)}
Given a real-valued feature space $\X = \R^d$ and a label space $\Y = [k] := \{1, \dots, k\}$ with $3 \leq k \in \N$ class labels, we denote a partially-labeled training dataset with $\D = \left\lbrace \left( \vc{x}_i, S_i \right) \mid i \in [n] \right\rbrace$ consisting of $n$ instances with features $\vc{x}_i \in \X$ and a non-empty set of candidate labels $S_i \subseteq \Y$ for all $i \in [n]$.
The ground-truth label $y_i \in \Y$ of instance $i$ is unknown.
However, one assumes $y_i \in S_i$ \citep{JinG02,LiuD12,LvXF0GS20}.
The goal is to train a classifier $g : \X \to \Y$ that minimizes the empirical loss with weak supervision only, that is, the exact ground truth labels $y_i$ are unavailable during training.
We use label weight vectors $\vc{\lbl}_{i} \in [0, 1]^k$ with $\| \vc{\lbl}_i \|_1 = 1$ to represent the model's knowledge about the labeling of instance $i \in [n]$.
Thereby, $\| \vc{\lbl}_i \|_p = (\sum_{j = 1}^{k} |{\lbl}_{ij}|^p)^{1/p}$ denotes the $p$-norm.
${\lbl}_{ij} \in [0, 1]$ denotes class $j$'s weight regarding instance $i$.
We typeset vectors in bold font.

\subsection{Subjective Logic (SL)}
Inspired by Dempster-Shafer theory \citep{Dempster08a,Shafer86}, \citet{Josang16} proposes a theory of evidence, called subjective logic, that explicitly represents (epistemic) uncertainty in prediction-making.
In subjective logic, the tuple $\omega_i = (\vc{\sublog{b}}_i, \vc{\sublog{a}}_i, \sublog{u}_i) \in [0, 1]^k \times [0, 1]^k \times [0, 1]$ denotes a multinomial opinion about instance $i \in [n]$ with $\vc{\sublog{b}}_i$ representing the belief mass of the class labels, $\vc{\sublog{a}}_i$ representing the prior knowledge about the class labels, and ${\sublog{u}}_i$ explicitly represents the uncertainty involved in predicting $i$.
$\omega_i$ satisfies $\| \vc{\sublog{a}}_i \|_1 = 1$ and requires additivity, that is, ${\sublog{u}}_i + \| \vc{\sublog{b}}_{i} \|_1 = 1$ for all $i \in [n]$.
The projected probability is defined by $\bar{\vc{p}}_{i} = \vc{\sublog{b}}_{i} + \vc{\sublog{a}}_{i} {\sublog{u}}_i \in [0, 1]^k$ for $i \in [n]$.
$\bar{\vc{p}}_{i}$ induces a probability measure $\mathbb{P}_i$ on the measurable space $(\Y, 2^{\Y})$ with $\mathbb{P}_i(A) = \sum_{j \in A} \bar{p}_{ij}$ for all $A \in 2^{\Y}$.
Given instance $i \in [n]$, features $\vc{x}_i \in \X$, and a prediction model $f : \R^{d} \to \R^{k}_{\geq 0}$, we set $\vc{\sublog{b}}_i = f(\vc{x}_i) / (k + \| f(\vc{x}_i) \|_1)$, $\vc{\sublog{a}}_i$ using prior knowledge, and ${\sublog{u}}_i = k / (k + \| f(\vc{x}_i) \|_1)$.
The multinomial opinion $\omega_i$ can be expressed in terms of a Dirichlet-distributed random variable with parameters $\vc{\alpha}_i = f(\vc{x}_i) + 1$, that is,
$\expect_{\vc{p}_{i} \sim \dirich(\vc{\alpha}_i)}[\vc{p}_{i}] := \vc{\alpha}_{i} / \| \vc{\alpha}_{i} \|_1 = \vc{\sublog{b}}_{i} + \vc{\sublog{a}}_{i} {\sublog{u}}_i = \bar{\vc{p}}_{i}$,
with $\vc{\sublog{b}}_i$ and $\sublog{u}_i$ as defined above, and uniform prior $\sublog{a}_{ij} = 1/k$ ($i \in [n]$, $j \in [k]$).
A multinomial opinion $\omega_i$ that is maximally uncertain, that is, $\sublog{u}_i = 1$, defaults to prior knowledge $\vc{\sublog{a}}_i$: In this case, $\bar{\vc{p}}_{i} = \vc{\sublog{a}}_{i}$ for $i \in [n]$.
In the following, we provide an example of multinomial opinions in subjective logic.
\vspace{0.5em}
\begin{example}
    \label{ex:sl}
    Let $n = 2$, $k = 3$, $\Y = \{1, 2, 3\}$, $\vc{\sublog{b}}_1 = (2/3, 1/6, 1/6)$, $\vc{\sublog{b}}_2 = (1/2, 0, 0)$, $\vc{\sublog{a}}_1 = \vc{\sublog{a}}_2 = (1/3, 1/3, 1/3)$, ${\sublog{u}}_1 = 0$, and ${\sublog{u}}_2 = 1/2$.
    Then, both multinomial opinions, $\omega_1 = (\vc{\sublog{b}}_1, \vc{\sublog{a}}_1, \sublog{u}_1)$ and $\omega_2 = (\vc{\sublog{b}}_2, \vc{\sublog{a}}_2, \sublog{u}_2)$, yield the same projected probabilities $\bar{\vc{p}}_{1} = \bar{\vc{p}}_{2} = (2/3, 1/6, 1/6)$.
    While $\omega_1$ and $\omega_2$ both induce the same probability measure on $(\Y, 2^{\Y})$, $\omega_2$ contains more uncertainty than $\omega_1$, that is, there is more evidence that supports $\omega_1$ than $\omega_2$.
    Also, both multinomial opinions induce a Dirichlet-distributed random variable with equal mean but different variances indicating different degrees of certainty about the labeling, which will be helpful in disambiguating partially-labeled data.
\end{example}

\section{Our Method: RobustPLL}
\label{sec:main}
We present a novel PLL method that yields robust predictions in terms of good predictive performance, robustness against out-of-distribution examples, and robustness against adversarial examples.
Tackling all is especially challenging in PLL as full ground truth is not available.

Given a prediction, one commonly uses softmax normalization, $\sm : \R^k \to [0, 1]^k$,  to output a discrete probability distribution over possible targets \citep{Bishop07}.
It has been noted, however, that softmax normalization cannot represent the uncertainty involved in prediction-making \citep{HullermeierW21,SaleCH23}, which is also evident from Example~\ref{ex:sl}, where different amounts of uncertainty can be associated with the same probability measure.
In partial-label learning, where candidate labels are iteratively refined, it is crucial to accurately reflect the uncertainty involved in the candidate labels to effectively propagate labeling information.
Our method explicitly represents uncertainty through the SL framework by using a neural-network that parameterizes a Dirichlet distribution rather than using softmax normalization.
By jointly learning the candidate label weights as well as their associated uncertainty, our approach builds robust representations, which help in dealing with out-of-distribution data and adversarially corrupted features.

Similar to an expectation-maximization procedure, we interleave learning the parameters of our prediction model with updating the current labeling information of all instances based on the discovered knowledge and prior information.

Algorithm~\ref{alg:robust-pll} outlines our method: \textsc{RobustPll}.
First, we initialize the label weights $\lbl_{ij}$ in Line~1.
These weights $\lbl_{ij}$ represent the degree to which instance $i$ has the correct label $j$.
Section~\ref{sec:init-weights} discusses their initialization and interpretation.
In Line~2, we set up our model $f : \R^d \to \R^k_{\geq 0}$ and its parameters $\vc{\theta}$.
Our framework is independent of the concrete model choice.
For example, one may use MLPs \citep{rumelhart1986learning}, LeNet \citep{LeCunBBH98}, or ResNet \citep{HeZRS16}.
One only needs to modify the last layer, which is required to be a ReLU layer to enforce non-negative outputs for the SL framework.
Lines~3--8 contain the main training loop of our approach.
We train for a total of $T$ epochs.
Note that, in practice, we make use of mini-batches.
We set the annealing coefficient $\lambda_t$ in Line~4.
The coefficient controls the influence of the regularization term in $\hat{\risk}(f; \lambda_t)$, which is discussed in Section~\ref{sec:regularization}.
In Line~5, we then compute the empirical risk $\hat{\risk}(f; \lambda_t)$ and update the model parameters $\vc{\theta}$ in Line~6.
Those steps are discussed in Section~\ref{sec:training}.
Thereafter, we update the label weights $\lbl_{ij}$ in Line~7 as shown in Section~\ref{sec:weight-update}.

The remainder of Section~\ref{sec:main} presents further analyses.
In Section~\ref{sec:reinterpret}, we discuss our reinterpretation of the label weight update within SL.
Section~\ref{sec:bound} bounds the rate of change of $\vc{\lbl}_{i}$ and Section~\ref{sec:cross-entropy} demonstrates why the squared error loss is superior to the cross-entropy loss in our setting.
Section~\ref{sec:runtime} discusses our approach's runtime.

\begin{algorithm}[t]
    \caption{\textsc{RobustPll} (Our proposed method)}
    \label{alg:robust-pll}

    \begin{algorithmic}[1]
        \item[\textbf{Input:}] Partially-labeled dataset $\D = \lbrace (\vc{x}_i, S_i) \in \X \times 2^{\Y} \mid i \in [ n ] \rbrace$ with $n$ instances, and $k = |\Y|$ classes;
        \item[\textbf{Output:}] Model parameters $\vc{\theta}$;
        \vspace{0.2em}
        \STATE Init weights $\lbl_{ij} \leftarrow \mathds{1}_{\lbrace j \in S_i \rbrace} / |S_i|$ for $i \in [n]$, $j \in [k]$;
        \vspace{0.2em}
        \STATE Init model $f$ and its parameters $\vc{\theta}$;
        \vspace{0.2em}
        \FOR{$t = 1, \dots, T$}
        \vspace{0.2em}
        \STATE $\lambda_t \leftarrow \min(2t / T, 1)$
        \STATE Compute the empirical risk $\hat{\risk}(f; \lambda_t)$ by~\eqref{eq:risk};
        \vspace{0.2em}
        \STATE Update $\vc{\theta}$ by backpropagation;
        \vspace{0.2em}
        \STATE Set label weights $\vc{\lbl}_i$ ($i \in [n]$) to the solution $\vc{\lbl}_i^{*}$ of~\eqref{eq:opt-label-weights} using the closed-form solution in Proposition~\ref{prop:optimal};
        \vspace{0.2em}
        \ENDFOR
    \end{algorithmic}
\end{algorithm}

\subsection{Initializing the Label Weights}
\label{sec:init-weights}
The label weights $\lbl_{ij}$ represent the current knowledge of our method about instance $i$ having the correct label $j$.
They must sum to one, that is, $\| \vc{\lbl}_i \|_1 = 1$, and must be zero if $j$ is not a candidate label of instance $i$, that is, $\lbl_{ij} = 0$ if $j \notin S_i$ ($i \in [n]$).
We initialize the label weights with
\begin{equation}
    \lbl_{ij} = \mathds{1}_{\lbrace j \in S_i \rbrace} / |S_i| \text{,} \label{eq:init}
\end{equation}
where $\mathds{1}_{\lbrace \cdot \rbrace}$ denotes the indicator function.
$\lbl_{ij}$ satisfies both requirements.
\eqref{eq:init} can be written as a multinomial opinion $\omega_i = (\vc{\sublog{b}}_i, \vc{\sublog{a}}_i, \sublog{u}_i)$ in SL with maximal uncertainty, that is, $\vc{\lbl}_i = \vc{\sublog{b}}_i + \vc{\sublog{a}}_{i} \sublog{u}_{i}$ with zero belief $\sublog{b}_{ij} = 0$, uniform prior weights $\sublog{a}_{ij} = \mathds{1}_{\lbrace j \in S_i \rbrace} / |S_i|$, and maximal uncertainty $\sublog{u}_{i} = 1$ ($i \in [n]$, $j \in [k]$).
Note that we have $\vc{\lbl}_i = \vc{\sublog{a}}_{i}$ at initialization: The label weights $\vc{\lbl}_i$ are solely determined by prior knowledge about the candidate sets encoded in $\vc{\sublog{a}}_{i}$.

\subsection{Training a Model}
\label{sec:training}
We interleave learning the parameters $\vc{\theta}$ of a model $f : \R^d \to \R^k_{\geq 0}$ (Lines~4--6) with updating the label weights $\vc{\lbl}_i$ based on the discovered knowledge (Line~7).
Our model $f$ does not directly output discrete probabilities (for example, via softmax) as a single probability mass function cannot reflect the degree of uncertainty involved in prediction-making, which we illustrate in Example~\ref{ex:model}.
Instead, the model $f$ outputs evidence supporting a particular class label, which parameterizes a Dirichlet distribution $\dirich(\vc{\alpha}_i)$ with
\begin{align}
    \vc{\alpha}_i = f(\vc{x}_i; \vc{\theta}) + 1 \in \R^{k}_{\geq 1} \quad\text{for}\quad i \in [n] \text{.} \label{eq:alpha}
\end{align}

To fit $f$ to the label weights $\vc{\lbl}_{i}$, we use a loss formulation similar to \citet{SensoyKK18}.
The loss regarding a fixed instance $i$ is characterized by the expected value of the squared distance of $\vc{\lbl}_i$ and $\vc{p}_i \sim \dirich(\vc{\alpha}_i)$ with $\vc{\alpha}_i$ as in~\eqref{eq:alpha}.
For an instance $i \in [n]$ with features $\vc{x}_i \in \X$ and label weights $\vc{\lbl}_i \in [0, 1]^k$, the squared error using the bias-variance decomposition is
\begin{align}
    \loss(f(\vc{x}_i; \vc{\theta}), \vc{\lbl}_i) & = \expect_{\vc{p}_i \sim \dirich(\vc{\alpha}_i)} \| \vc{\lbl}_i - \vc{p}_i \|_{2}^{2} \label{eq:loss}                                                                                                                                                                                  \\
                                                 & = \sum_{j=1}^{k} \expect[ (\lbl_{ij} - p_{ij})^2 ]                                                                                                                                                                                                                                     \\
                                                 & \overset{(i)}{=} \sum_{j=1}^{k} \big[ (\lbl_{ij} - \expect[p_{ij}] )^2 \!+\! \var[ p_{ij} ] \big]                                                                                                                                                                                      \\
                                                 & \overset{(ii)}{=} \sum_{j=1}^{k} \Big[ \underbrace{(\lbl_{ij} - \bar{p}_{ij})^2}_{=: \loss_{ij}^{\operatorname{err}}} + \underbrace{\frac{\bar{p}_{ij} (1 - \bar{p}_{ij})}{1 + \| \vc{\alpha}_i \|_1}}_{=: \loss_{ij}^{\operatorname{var}}} \Big] \text{,}\quad \label{eq:loss-decomp}
\end{align}
with $\bar{\vc{p}}_i = \expect_{\vc{p}_i \sim \dirich(\vc{\alpha}_i)} [\vc{p}_i] = \vc{\alpha}_i / \| \vc{\alpha}_i \|_1$.
$(i)$ holds by expansion of the squared term and rearrangement and $(ii)$ by the known variance of Dirichlet random variables.
Fortunately, one does not need to numerically approximate the integral of the expected value in~\eqref{eq:loss}.
One can directly compute $\loss$ using the outputs of $f$ only.
In the following, we give an example highlighting the differences to softmax normalization.
\vspace{0.5em}
\begin{example}
    \label{ex:model}
    Let $n = 2$, $k = 3$, $\Y = \{1, 2, 3\}$, $\vc{x}_1, \vc{x}_2 \in \X$, $f(\vc{x}_1) = (4, 1, 1)$, and $f(\vc{x}_2) = (7, 4, 4)$.
    Using softmax normalization, both predictions yield the same discrete probabilities, that is, $\sm(f(\vc{x}_1)) = \sm(f(\vc{x}_2)) \approx (0.910, 0.045, 0.045)$, although $\vc{x}_1$ and $\vc{x}_2$ have different activation values.
    In our setting, we have $\vc{\alpha}_1 = f(\vc{x}_1) + 1 = (5, 2, 2)$ and $\vc{\alpha}_2 = f(\vc{x}_2) + 1 = (8, 5, 5)$ by~\eqref{eq:alpha}, and the predicted probabilities $\bar{\vc{p}}_1 = \expect_{\vc{p}_1 \sim \dirich(\vc{\alpha}_1)} [\vc{p}_1] = (5/9, 2/9, 2/9)$ and $\bar{\vc{p}}_2 = \expect_{\vc{p}_2 \sim \dirich(\vc{\alpha}_2)} [\vc{p}_2] = (4/9, 5/18, 5/18)$.
    While both probabilities are still close, similar to the softmax normalization, the different variances of the Dirichlet distributions represent different degrees of uncertainty, that is, $\operatorname{Var}_{\vc{p}_1 \sim \dirich(\vc{\alpha}_1)} [\vc{p}_1] \approx (0.025, 0.017, 0.017)$ and $\operatorname{Var}_{\vc{p}_2 \sim \dirich(\vc{\alpha}_2)} [\vc{p}_2] \approx (0.013, 0.011, 0.011)$.
    Since $f(\vc{x}_2)$ has higher activation values, there is less associated uncertainty across all classes.
    Hence, $\dirich(\vc{\alpha}_2)$ has less variance than $\dirich(\vc{\alpha}_1)$.
\end{example}
\vspace{0.5em}
The loss term $\loss$ in \eqref{eq:loss-decomp} can be separated into an error and variance component, $\loss_{ij}^{\operatorname{err}}$ and $\loss_{ij}^{\operatorname{var}}$, respectively.
$\loss_{ij}^{\operatorname{err}}$ enforces model fit and $\loss_{ij}^{\operatorname{var}}$ acts as regularization term and incentivizes the decrease of the variance of the Dirichlet distribution parameterized by $f$.
To prioritize model fit, it is desirable that $\loss_{ij}^{\operatorname{err}} > \loss_{ij}^{\operatorname{var}}$ if $\lbl_{ij}$ and $\bar{p}_{ij}$ deviate too much, which we discuss in the following.

\begin{proposition}
    \label{prop:var}
    Given instance $(\vc{x}_i, S_i) \in \D$, parameters $\vc{\theta}$, label weights $\vc{\lbl}_{i}$, and $\bar{\vc{p}}_{i} = \vc{\alpha}_{i} / \| \vc{\alpha}_{i} \|_1$, it holds that $\loss_{ij}^{\operatorname{err}} < \loss_{ij}^{\operatorname{var}}$ if and only if $\bar{p}_{ij} - \sqrt{\loss_{ij}^{\operatorname{var}}} < \lbl_{ij} < \bar{p}_{ij} + \sqrt{\loss_{ij}^{\operatorname{var}}}$, for all $i \in [n]$ and $j \in [k]$.
\end{proposition}

Proposition~\ref{prop:var} sheds light on the magnitudes of $\loss_{ij}^{\operatorname{err}}$ and $\loss_{ij}^{\operatorname{var}}$.
When $\lbl_{ij}$ is within one standard deviation from $\bar{p}_{ij}$, $\lbl_{ij}$ is close to $\bar{p}_{ij}$.
In this regime, reducing variance is more important than improving model fit, that is, $\loss_{ij}^{\operatorname{err}} < \loss_{ij}^{\operatorname{var}}$.
Reducing the variance of the Dirichlet distribution is equivalent to a reduction of the uncertainty about the prediction.
When $\lbl_{ij}$ is outside one standard deviation from $\bar{p}_{ij}$, model fit is more important, that is, $\loss_{ij}^{\operatorname{err}} > \loss_{ij}^{\operatorname{var}}$.
This property guarantees that we can jointly learn the candidate label weights as well as their associated uncertainty seamlessly, which helps in creating robust representations to deal with out-of-distribution data and adversarial modifications of the features.

\subsection{Regularization}
\label{sec:regularization}
Given an instance $\vc{x}_i \in \X$, the correct label $y_i \in \Y$ is hidden within $S_i \subseteq \Y$ (Section~\ref{sec:prelim-pll}).
Therefore, our model $f$ should not allocate any evidence to incorrect labels, that is, $f_j(\vc{x}_i; \vc{\theta})$ should be zero for $i \in [n]$ and $j \notin S_i$.
Similar to \citet{SensoyKK18}, we add a regularization term to the risk computation to avoid evidence supporting incorrect labels $j \notin S_i$.
Let $\tilde{\alpha}_{ij} = \alpha_{ij}$ if $j \notin S_i$, else $\tilde{\alpha}_{ij} = 1$, for $i \in [n]$, $j \in S_i$.
We then achieve maximal uncertainty about predicting $j \notin S_i$ by considering the KL-divergence between $\dirich( \tilde{\vc{\alpha}}_i )$ and $\dirich( 1 )$.
We compute the empirical risk as
\begin{align}
    \hat{\risk}(f; \lambda_t) = \frac{1}{n} \sum_{i=1}^{n} \big[ \loss(f(\vc{x}_i; \vc{\theta}), \vc{\lbl}_i) + \lambda_t \kl( \dirich( \tilde{\vc{\alpha}}_i ) \| \dirich( 1 ) ) \big]\text{.} \label{eq:risk}
\end{align}
This has a positive effect on classification as $f$ also learns from \emph{negative} examples, that is, $f$ should be maximally uncertain about predicting $j \notin S_i$.
However, to avoid our model $f$ from being uncertain about all labels, we gradually increase the regularization coefficient $\lambda_t$.
This regularization directly benefits the robustness of our method when dealing with out-of-distribution and adversarial data.
Given two Dirichlet-distributed random variables with parameters $\dirich( \tilde{\vc{\alpha}}_i )$ and $\dirich( 1 )$ respectively, their KL divergence permits a closed-form expression \citep{kldiv}.
One updates the parameters $\vc{\theta}$ by backpropagation of~\eqref{eq:risk}.
Note that the KL term also depends on the parameters $\vc{\theta}$.

\subsection{Updating the Label Weights}
\label{sec:weight-update}
After updating the parameters $\vc{\theta}$, we extract the learned knowledge about the class labels to iteratively disambiguate the candidate sets $S_i$.
For a fixed instance $(\vc{x}_i, S_i) \in \D$ and model parameters $\vc{\theta}$, we want to find the optimal label weights $\vc{\lbl}_i \in [0, 1]^k$ that minimize~\eqref{eq:loss-decomp}, while maintaining all prior information about the candidate set membership, that is, $\lbl_{ij} = 0$ if $j \notin S_i$.
We cannot directly assign $\bar{\vc{p}}_i$ to the label weights $\vc{\lbl}_i$ since $f$ can allocate evidence to incorrect labels, that is, $f_j(\vc{x}_i; \vc{\theta}) \geq 0$ for $j \notin S_i$.
In Line~7 of Algorithm~\ref{alg:robust-pll}, we assign $\vc{\lbl}_i \in [0, 1]^k$ to the solution of
\begin{align}
    \min_{\vc{\lbl}'_i \in [0, 1]^k} \loss(f(\vc{x}_i; \vc{\theta}), \vc{\lbl}'_i) \quad \text{subject to}\quad \|\vc{\lbl}'_i\|_1 = 1 \ \text{and}\ \lbl'_{ij} = 0 \ \text{if}\ j \notin S_i \text{.} \label{eq:opt-label-weights}
\end{align}
\eqref{eq:opt-label-weights} permits a closed-form solution, which is as follows.

\begin{proposition}
    \label{prop:optimal}
    Given a fixed instance $(\vc{x}_i, S_i) \in \D$, model parameters $\vc{\theta}$, and $\bar{\vc{p}}_i = \vc{\alpha}_i / \| \vc{\alpha}_i \|_1$, the optimization problem~\eqref{eq:opt-label-weights}, with $\loss$ as in~\eqref{eq:loss-decomp}, has the solution
    \begin{align}
        \lbl_{ij}^{*} = \begin{cases}
                            \bar{p}_{ij} + \frac{1}{|S_i|} (1 - \sum_{j' \in S_i} \bar{p}_{ij'} ) & \text{if $j \in S_i$,} \\
                            0                                                                     & \text{else.}           \\
                        \end{cases}
    \end{align}
\end{proposition}

The proof of Proposition~\ref{prop:optimal} (see Appendix~\ref{sec:proof-optimal}) first shows that $\lbl_{ij}^{*}$ is a feasible solution for the constraints in~\eqref{eq:opt-label-weights} and then establishes optimality using the Lagrangian multiplier method since $\loss$ is continuous and differentiable.
The solution $\lbl_{ij}^{*}$ uniformly re-distributes all weight of labels not in $S_i$ to labels, which are in $S_i$.
This guarantees a minimal loss.
Notably, the update strategy in Proposition~\ref{prop:optimal} differs from the heuristic update strategies proposed in related work \citep{LvXF0GS20,0009LLQG23,crosel2024}.

\subsection{Reinterpreting the Label Weights}
\label{sec:reinterpret}
Recall from Section~\ref{sec:prelim-pll} that subjective logic allows decomposing the projected probabilities $\bar{\vc{p}}_i$ into a multinomial opinion $\omega_i$ with a belief and uncertainty term, that is, $\bar{\vc{p}}_i = \vc{\sublog{b}}_{i} + \vc{\sublog{a}}_{i} {\sublog{u}}_i$.
This representation directly allows quantifying the uncertainty involved in prediction-making.
It is desirable that the label weight update $\vc{\lbl}_{i}^{*}$ in Section~\ref{sec:weight-update} can also be written as such a multinomial opinion to allow for the direct quantification of the involved uncertainty.

\begin{proposition}
    \label{prop:reinterpret}
    Given a fixed instance $(\vc{x}_i, S_i) \in \D$ and parameters $\vc{\theta}$, the solution $\lbl_{ij}^{*}$ of~\eqref{eq:opt-label-weights}, which is given by Proposition~\ref{prop:optimal}, is equivalent to $\vc{\lbl}_{i}^{*} = \vc{\sublog{b}}_{i} + \vc{\sublog{a}}_{i} \sublog{u}_{i}$ with
    \begin{align}
        \sublog{b}_{ij} = \mathds{1}_{\lbrace j \in S_i \rbrace} \frac{f_j(\vc{x}_i; \vc{\theta})}{\| \vc{\alpha}_i \|_1} \ \text{with}\  \vc{\alpha}_i = f(\vc{x}_i; \vc{\theta}) + 1 \text{, } \sublog{u}_{i} = 1 - \sum_{j \in S_i} \sublog{b}_{ij} \text{, and } \sublog{a}_{ij} = \frac{ \mathds{1}_{\lbrace j \in S_i \rbrace}} {|S_i|} \text{.} \notag
    \end{align}
\end{proposition}

Proposition~\ref{prop:reinterpret} allows reinterpreting $\vc{\lbl}_i^{*}$ (Proposition~\ref{prop:optimal}) as such a multinomial opinion $\omega_i$ about instance $i$.
Proposition~\ref{prop:reinterpret} establishes that the belief of our prediction model in non-candidate labels directly contributes to the uncertainty $\sublog{u}_i$ in predicting instance $i \in [n]$.
The uncertainty term $\sublog{u}_i$ arises from unallocated belief mass $\sublog{b}_{ij}$, that is, $1 - \sum_{j \in S_i} \sublog{b}_{ij}$ for $i \in [n]$.
Also, the prior weights $\sublog{a}_{ij}$ are defined similarly to the initial label weights $\lbl_{ij}$ (see Section~\ref{sec:init-weights}).
The prior weights are uniformly distributed among all candidate labels.
The result in Proposition~\ref{prop:reinterpret} establishes that our proposed update strategy (Proposition~\ref{prop:optimal}) is valid within subjective logic.

\subsection{Bounding the Label Weights}
\label{sec:bound}
This section examines how the model's probability outputs $\bar{\vc{p}}_i$ and the label weights $\vc{\lbl}_i$ interact with each other.
In the following, we provide an upper bound of the change of $\vc{\lbl}_i$'s values over time.
As the label weights $\vc{\lbl}_i$ are the prediction targets in~\eqref{eq:loss-decomp}, it is desirable that the $\vc{\lbl}_i$ do not oscillate, which we detail in the following.

\begin{proposition}
    \label{prop:bound}
    Given an instance $(\vc{x}_i, S_i) \in \D$, its label weights $\vc{\lbl}_i^{(t)} \in [0, 1]^k$, and the model's probability outputs $\bar{\vc{p}}_i^{(t)}$ at epoch $t \in \N$, we can give the following upper bound:
    \begin{align}
        0 \leq \| \vc{\lbl}_i^{(t + 1)} - \vc{\lbl}_i^{(t)} \|_{2}^{2} \leq \| \bar{\vc{p}}_i^{(t + 1)} - \bar{\vc{p}}_i^{(t)} \|_{2}^{2} \ \ \text{ for }\ i \in [n] \text{.}
    \end{align}
\end{proposition}

It indicates that the label weights $\vc{\lbl}_i$ change at most as fast as the model's probability outputs $\bar{\vc{p}}_i$ between consecutive epochs.
An immediate consequence is that the convergence of the model training and its probability outputs $\bar{\vc{p}}_i$, that is, $\| \bar{\vc{p}}_i^{(t+1)} - \bar{\vc{p}}_i^{(t)} \|_{2}^{2} \to 0$ for $t \to \infty$, implies the convergence of the label weight vectors $\vc{\lbl}_i$, that is, $\| \vc{\lbl}_i^{(t + 1)} - \vc{\lbl}_i^{(t)} \|_{2}^{2} \to 0$, which are extracted from the model.
This property is desirable as it shows that the label weights $\vc{\lbl}_i$ do not oscillate if the model's probability outputs $\bar{\vc{p}}_i$, which depend on the model parameters $\vc{\theta}$, converge.

\subsection{Cross-Entropy Loss}
\label{sec:cross-entropy}
Although we use the squared error loss~\eqref{eq:loss}, it is worth considering the commonly used cross-entropy loss.
Given an instance $\vc{x}_i \in \X$, a model $f$ with parameters $\vc{\theta}$, and label weights $\vc{\lbl}_i$, a cross-entropy formalization similar to~\citet{SensoyKK18} is given by
\begin{align}
    \loss_{\operatorname{CE}}(f(\vc{x}_i; \vc{\theta}), \vc{\lbl}_i)
    = \expect [ -\sum_{j=1}^{k} \lbl_{ij} \log p_{ij} ]
    \overset{(i)}{=} \vc{\lbl}_{i} \cdot \vc{\Psi}_{i}
    \text{,}\quad \label{eq:ce}
\end{align}
with $\Psi_{ij} = \psi(\| \vc{\alpha}_i \|_1) - \psi(\alpha_{ij})$ and $\psi$ denoting the di\-gamma function.
$(i)$ holds because of the linearity of the expected value and $\expect_{p_{ij} \sim \dirich_j(\vc{\alpha}_i)} \log p_{ij} = \psi(\alpha_{ij}) - \psi(\| \vc{\alpha}_i \|_1)$.
In the following, we establish the optimal choice of $\vc{\lbl}_i$ in our optimization problem~\eqref{eq:opt-label-weights} using the cross-entropy loss~\eqref{eq:ce}.

\begin{proposition}
    \label{prop:optimal-ce}
    Given a fixed instance $(\vc{x}_i, S_i) \in \D$ and parameters $\vc{\theta}$, optimization problem~\eqref{eq:opt-label-weights}, using the cross-entropy loss~\eqref{eq:ce}, has the closed-form solution
    \begin{align}
        \lbl_{ij}^{*} = \begin{cases}
                            1 & \text{if $j = \arg\min_{j' \in S_i} \Psi_{ij'}$,} \\
                            0 & \text{else.}                                      \\
                        \end{cases}
    \end{align}
\end{proposition}

This suggests that the cross-entropy loss~\eqref{eq:ce} enforces a quite aggressive label-weight update strategy setting all mass on one class label.
Also, the label weights given by Proposition~\ref{prop:optimal-ce} cannot be reinterpreted in subjective logic as discussed in Section~\ref{sec:reinterpret}.
The squared error loss also performs better than the cross-entropy loss empirically.
For these reasons, all consecutive experiments are conducted using the update strategy in Proposition~\ref{prop:optimal}.

\subsection{Runtime Analysis}
\label{sec:runtime}
Recall from Section~\ref{sec:training} that one does not need to numerically approximate the integral within the expectation value in~\eqref{eq:loss}.
Given label weights $\vc{\lbl}_i$, one can directly compute $\loss$ using the outputs of $f$ only.
The computation of the KL divergence between two Dirichlet-distributed random variables with parameters $\dirich( \tilde{\vc{\alpha}}_i )$ and $\dirich( 1 )$, respectively, admits a closed-form expression \citep{kldiv}, leading to an overall linear runtime in $n$ to compute $\hat{\risk}(f; \lambda_t)$ (Algorithm~\ref{alg:robust-pll}, Line~5).
In Line~7 of Algorithm~\ref{alg:robust-pll}, Proposition~\ref{prop:optimal} also permits updating $\lbl_{ij}$ in linear time regarding $n$.
Therefore, our method's runtime is dominated solely by the forward and backward pass of the employed model $f$.

\section{Experiments}
\label{sec:exp}
Section~\ref{sec:competitors} summarizes all methods that we compare against and Section~\ref{sec:setup} outlines the experimental setup.
Thereafter, Section~\ref{sec:pred} analyzes the methods' robustness against PLL noise, Section~\ref{sec:ood} against out-of-distribution samples, and Section~\ref{sec:adv} against adversarial perturbations.

\subsection{Algorithms for Comparison}
\label{sec:competitors}
There are many PLL algorithms from which we pick the best-performing and commonly used ones for comparison.
We cover classic algorithms and deep-learning techniques and complement these methods with strong baselines.

We consider 13 methods: \textsc{PlKnn} \citep{HullermeierB06}, \textsc{PlSvm} \citep{NguyenC08}, \textsc{Ipal} \citep{ZhangY15a}, \textsc{PlEcoc} \citep{ZhangYT17}, \textsc{Proden} \citep{LvXF0GS20}, \textsc{Rc} \citep{FengL0X0G0S20}, \textsc{Cc} \citep{FengL0X0G0S20}, \textsc{Valen} \citep{XuQGZ21}, \textsc{Cavl} \citep{ZhangF0L0QS22}, \textsc{Pop} \citep{0009LLQG23}, \textsc{CroSel} \citep{crosel2024}, \textsc{DstPll} \citep{fuchs2024partiallabel}, and \textsc{RobustPll} (our method).

Additionally, we benchmark various extensions known to obtain robust results in the supervised domain: \textsc{Proden} with L2-regularization (\textsc{Proden+L2}), \textsc{Proden} with dropout\footnote{Dropout is also applied in testing to form an explicit ensemble.} (\textsc{Proden+Dropout}; \citealt{SrivastavaHKSS14}), \textsc{Proden} for disambiguating the partial labels and then training an evidential-deep-learning classifier \citep{SensoyKK18} in a supervised manner (\textsc{Proden+Edl}), an ensemble of 5 \textsc{Proden} classifiers (\textsc{Proden+Ens}; \citealt{Lakshminarayanan17}), an ensemble of 5 \textsc{Proden} classifiers trained on adversarial examples (\textsc{Proden+AdvEns}; \citealt{Lakshminarayanan17}), and an ensemble of our method (\textsc{RobustPll+Ens}).

For a fair comparison, we use the same base model, that is, a $d$-300-300-300-$k$ MLP \citep{werbos1974beyond}, for all neural-network-based approaches.
Appendix~\ref{sec:bonus-exp} discusses the specific hyperparameter choices of all approaches in more detail.
We publicly provide all code and data for reproducibility.\textsuperscript{\ref{fnt:code}}

\subsection{Experimental Setup}
\label{sec:setup}
As is common in the literature \citep{ZhangYT17,0009LLQG23}, we conduct experiments on supervised datasets with added noise as well as on real-world partially-labeled datasets.
We use four supervised MNIST-like datasets with added noise and six real-world PLL datasets.
For the supervised datasets, we use MNIST \citep{uci-mnist}, KMNIST \citep{uci-kmnist}, FMNIST \citep{uci-fmnist}, and NotMNIST \citep{notmnist}.
For the real-world datasets, we use \emph{bird-song} \citep{BriggsFR12}, \emph{lost} \citep{CourST11}, \emph{mir-flickr} \citep{HuiskesL08}, \emph{msrc-v2} \citep{LiuD12}, \emph{soccer} \citep{ZengXJCGXM13}, and \emph{yahoo-news} \citep{GuillauminVS10}.

We use instance-dependent noise to introduce partial labels into the supervised datasets \citep{ZhangZW0021}.
This strategy first trains a supervised classifier $g : \mathbb{R}^d \to [0, 1]^k$, which outputs probabilities $g_j(\vc{x}_i)$ for instance $i \in [n]$ and class labels $j \in [k]$.
Given an instance's features $\vc{x}_i \in \X$ with correct label $y_i \in \Y$, a flipping probability of $\xi_j(\vc{x}_i) = g_j(\vc{x}_i) / \max_{j' \in \Y \setminus \lbrace y_i \rbrace} g_{j'}(\vc{x}_i)$ determines whether to add the incorrect label $j \neq y_i$ to the candidate set $S_i$.
Additionally, one divides $\xi_j(\vc{x}_i)$ by the mean probability $\frac{1}{k-1} \sum_{j' \neq y_i} \xi_{j'}(\vc{x}_i)$ of incorrect labels \citep{XuQGZ21,0009LLQG23}, which makes all labels more likely to appear.
While all MNIST-like datasets have ten class labels, the averages ($\pm$ std.) of the candidate set cardinalities are \SI{6.08}{} ($\pm$ 0.06) for MNIST, \SI{5.44}{} ($\pm$ 0.05) for FMNIST, \SI{5.73}{} ($\pm$ 0.04) for KMNIST, and \SI{6.99}{} ($\pm$ 0.09) for NotMNIST.
We publicly provide all code and data for reproducibility.\textsuperscript{\ref{fnt:code}}

\subsection{Robustness under PLL Noise}
\label{sec:pred}

\begin{table}[tp]
    \caption{
        Average test-set accuracies ($\pm$ std.) on the MNIST-like and real-world datasets.
        All experiments are repeated five times with different seeds to report mean and standard deviations.
        The MNIST-like datasets have added instance-dependent noise as discussed in Section~\ref{sec:setup}.
        The column for the real-world datasets contains averages across all six real-world datasets.
        We emphasize the best algorithm per dataset, as well as non-significant differences, using a student t-test with level $\alpha = 0.05$.
        We consider non-ensemble and ensemble methods separately.
        The triangles indicate our proposed methods.
    }
    \label{tab:accuracies}
    \begin{center}
        \begin{small}
            \begin{tabular}{
                >{\raggedright\arraybackslash}m{0.217\textwidth}
                >{\centering\arraybackslash}m{0.137\textwidth}
                >{\centering\arraybackslash}m{0.137\textwidth}
                >{\centering\arraybackslash}m{0.137\textwidth}
                >{\centering\arraybackslash}m{0.137\textwidth}
                >{\centering\arraybackslash}m{0.138\textwidth}
                }
                \toprule
                \multirow[c]{2}{*}[-0.075cm]{All methods} & \multicolumn{4}{c}{MNIST-like datasets with inst.-dep. noise} & \multirow[c]{2}{*}[-0.06cm]{\hspace{-0.16cm} \parbox{0.12\textwidth}{\centering Real-world datasets}}                                                                                     \\
                \cmidrule(lr){2-5}
                                                          & {\centering MNIST}                                            & {\centering FMNIST}                                                                                   & {\centering KMNIST}       & {\centering NotMNIST}     &                           \\
                \midrule
                \textsc{PlKnn} (2005)                     & 46.6 ($\pm$ 0.5)                                              & 41.9 ($\pm$ 0.4)                                                                                      & 52.2 ($\pm$ 0.4)          & 31.3 ($\pm$ 0.9)          & 50.3 ($\pm$ 8.8)          \\
                \textsc{PlSvm} (2008)                     & 32.4 ($\pm$ 5.0)                                              & 37.3 ($\pm$ 1.9)                                                                                      & 31.6 ($\pm$ 4.2)          & 39.2 ($\pm$ 3.9)          & 40.7 ($\pm$ 10.1)         \\
                \textsc{Ipal} (2015)                      & \textbf{96.0} ($\pm$ 0.4)                                     & 75.1 ($\pm$ 0.7)                                                                                      & \textbf{80.8} ($\pm$ 0.9) & 61.5 ($\pm$ 1.6)          & 57.8 ($\pm$ 7.1)          \\
                \textsc{PlEcoc} (2017)                    & 61.6 ($\pm$ 2.9)                                              & 49.6 ($\pm$ 4.5)                                                                                      & 40.6 ($\pm$ 2.7)          & 39.8 ($\pm$ 6.1)          & 42.7 ($\pm$ 19.8)         \\
                \textsc{Proden} (2020)                    & 93.2 ($\pm$ 0.5)                                              & \textbf{77.8} ($\pm$ 2.5)                                                                             & 76.6 ($\pm$ 0.5)          & 84.6 ($\pm$ 1.3)          & \textbf{64.1} ($\pm$ 7.4) \\
                \textsc{Proden+L2}                        & 93.3 ($\pm$ 0.4)                                              & \textbf{78.1} ($\pm$ 1.7)                                                                             & 76.4 ($\pm$ 0.6)          & 84.6 ($\pm$ 1.2)          & \textbf{64.1} ($\pm$ 7.5) \\
                \textsc{Proden+Edl}                       & 92.0 ($\pm$ 0.5)                                              & 74.9 ($\pm$ 2.4)                                                                                      & 74.5 ($\pm$ 0.7)          & 80.8 ($\pm$ 0.5)          & 49.6 ($\pm$ 21.0)         \\
                \textsc{Rc} (2020)                        & 93.0 ($\pm$ 0.4)                                              & \textbf{78.0} ($\pm$ 2.3)                                                                             & 76.5 ($\pm$ 0.7)          & 84.1 ($\pm$ 1.6)          & 62.1 ($\pm$ 8.9)          \\
                \textsc{Cc} (2020)                        & 93.1 ($\pm$ 0.2)                                              & 78.9 ($\pm$ 0.9)                                                                                      & 77.5 ($\pm$ 0.8)          & 83.5 ($\pm$ 0.9)          & 43.8 ($\pm$ 31.2)         \\
                \textsc{Valen} (2021)                     & 50.3 ($\pm$ 5.3)                                              & 59.6 ($\pm$ 1.9)                                                                                      & 37.3 ($\pm$ 1.3)          & 50.3 ($\pm$ 2.4)          & 53.8 ($\pm$ 9.0)          \\
                \textsc{Cavl} (2022)                      & 79.5 ($\pm$ 6.4)                                              & 72.9 ($\pm$ 2.4)                                                                                      & 64.6 ($\pm$ 6.5)          & 61.5 ($\pm$ 6.8)          & 61.4 ($\pm$ 6.7)          \\
                \textsc{Pop} (2023)                       & 92.5 ($\pm$ 0.6)                                              & \textbf{79.0} ($\pm$ 1.6)                                                                             & 77.6 ($\pm$ 0.2)          & 84.5 ($\pm$ 1.8)          & \textbf{64.1} ($\pm$ 7.5) \\
                \textsc{CroSel} (2024)                    & 95.3 ($\pm$ 0.1)                                              & \textbf{79.6} ($\pm$ 0.9)                                                                             & 79.6 ($\pm$ 0.6)          & \textbf{86.6} ($\pm$ 0.7) & 41.9 ($\pm$ 30.1)         \\
                \textsc{DstPll} (2024)                    & 62.2 ($\pm$ 0.9)                                              & 50.3 ($\pm$ 1.0)                                                                                      & 68.4 ($\pm$ 1.0)          & 38.2 ($\pm$ 0.7)          & 48.5 ($\pm$ 9.6)          \\
                $\blacktriangleright$ \textsc{RobustPll}  & \textbf{96.0} ($\pm$ 0.1)                                     & \textbf{79.6} ($\pm$ 3.0)                                                                             & \textbf{81.7} ($\pm$ 0.3) & \textbf{83.7} ($\pm$ 1.9) & 59.5 ($\pm$ 6.8)          \\
                \midrule
                \textsc{Proden+Dropout}                   & 92.5 ($\pm$ 0.6)                                              & 72.7 ($\pm$ 2.8)                                                                                      & 72.1 ($\pm$ 1.1)          & 78.0 ($\pm$ 2.5)          & 65.0 ($\pm$ 8.1)          \\
                \textsc{Proden+Ens}                       & 93.7 ($\pm$ 0.2)                                              & 78.0 ($\pm$ 2.3)                                                                                      & 77.3 ($\pm$ 0.5)          & \textbf{85.6} ($\pm$ 0.7) & 65.8 ($\pm$ 8.3)          \\
                \textsc{Proden+AdvEns}                    & 95.3 ($\pm$ 0.6)                                              & 77.9 ($\pm$ 2.3)                                                                                      & 77.7 ($\pm$ 0.9)          & \textbf{84.3} ($\pm$ 1.5) & \textbf{66.7} ($\pm$ 9.0) \\
                $\blacktriangleright$ \textsc{RPll+Ens}   & \textbf{96.3} ($\pm$ 0.1)                                     & \textbf{80.4} ($\pm$ 2.3)                                                                             & \textbf{82.9} ($\pm$ 0.5) & \textbf{85.9} ($\pm$ 1.6) & 63.6 ($\pm$ 7.8)          \\
                \bottomrule
            \end{tabular}
        \end{small}
    \end{center}
\end{table}

Robust PLL algorithms should exhibit good predictive performance when confronted with PLL noise from ambiguous candidate sets.
Table~\ref{tab:accuracies} shows the accuracies of all methods on the MNIST-like and real-world datasets.
All supervised datasets have added noise (see Section~\ref{sec:setup}).
We repeat all experiments five times to report averages and standard deviations.
The best algorithm per dataset, as well as algorithms with non-significant differences, are emphasized.
Thereby, we consider non-ensemble methods (top) and ensemble methods (bottom) separately for fairness.
We use a paired student t-test with level $\alpha = 0.05$ to test for significance.

Our method (\textsc{RobustPll}) performs best on the four MNIST-like datasets and comparably on the real-world datasets.
We observe a similar behavior regarding our ensemble method (\textsc{RobustPll+Ens}).
Our non-ensemble method even achieves comparable performance to the ensemble methods on the MNIST-like datasets.
Summing up, we perform most consistently well under high PLL noise levels.

\subsection{Out-of-distribution Robustness}
\label{sec:ood}

\begin{table}[tp]
    \caption{
        The difference in the predictive entropies on the test and OOD sets.
        The models have been trained on the MNIST train dataset with added noise as discussed in Section~\ref{sec:setup}.
        We report the area between the empirical CDFs, the KS statistic, and the maximum-mean discrepancy using the RBF kernel.
        A value of one is optimal.
        Also compare Figure~\ref{fig:ood} for a graphical representation.
        Negative values indicate that the predictions on the out-of-distribution set are taken more confidently than the predictions on the test set.
    }
    \label{tab:ood}
    \begin{center}
        \begin{small}
            \begin{tabular}{
                >{\raggedright\arraybackslash}m{0.3\textwidth}
                >{\centering\arraybackslash}m{0.21\textwidth}
                >{\centering\arraybackslash}m{0.21\textwidth}
                >{\centering\arraybackslash}m{0.21\textwidth}
                }
                \toprule
                \multirow[c]{2}{*}[-0.075cm]{All methods}    & \multicolumn{3}{c}{Difference in entropy on MNIST and NotMNIST}                                                           \\
                \cmidrule(ll){2-4}
                                                             & {\centering CDF Area}                                           & {\centering KS stat.}      & {\centering MMD}           \\
                \midrule
                \textsc{PlKnn} (2005)                        & \phantom{-}0.0172                                               & \phantom{-}0.1587          & \phantom{-}0.0429          \\
                \textsc{PlSvm} (2008)                        & \phantom{-}0.0114                                               & \phantom{-}0.2944          & \phantom{-}0.0296          \\
                \textsc{Ipal} (2015)                         & \phantom{-}0.0896                                               & \phantom{-}0.3665          & \phantom{-}0.1285          \\
                \textsc{PlEcoc} (2017)                       & -0.0216                                                         & -0.3674                    & -0.0544                    \\
                \textsc{Proden} (2020)                       & \phantom{-}0.1769                                               & \phantom{-}0.6550          & \phantom{-}0.4240          \\
                \textsc{Proden+L2}                           & \phantom{-}0.1853                                               & \phantom{-}0.6844          & \phantom{-}0.4410          \\
                \textsc{Proden+Edl}                          & \phantom{-}\textbf{0.4379}                                      & \phantom{-}0.7171          & \phantom{-}\textbf{0.6714} \\
                \textsc{Rc} (2020)                           & \phantom{-}0.1402                                               & \phantom{-}0.5560          & \phantom{-}0.3495          \\
                \textsc{Cc} (2020)                           & \phantom{-}0.0607                                               & \phantom{-}0.5587          & \phantom{-}0.1378          \\
                \textsc{Valen} (2021)                        & -0.7668                                                         & -0.9434                    & -1.2137                    \\
                \textsc{Cavl} (2022)                         & \phantom{-}0.0087                                               & \phantom{-}0.1555          & \phantom{-}0.0205          \\
                \textsc{Pop} (2023)                          & \phantom{-}0.1345                                               & \phantom{-}0.5570          & \phantom{-}0.3361          \\
                \textsc{CroSel} (2024)                       & \phantom{-}0.2278                                               & \phantom{-}\textbf{0.8360} & \phantom{-}0.5202          \\
                \textsc{DstPll} (2024)                       & \phantom{-}0.1723                                               & \phantom{-}0.5097          & \phantom{-}0.3243          \\
                $\blacktriangleright$ \textsc{RobustPll}     & \phantom{-}\textbf{0.3855}                                      & \phantom{-}0.7345          & \phantom{-}\textbf{0.6707} \\
                \midrule
                \textsc{Proden+Dropout}                      & \phantom{-}0.2541                                               & \phantom{-}0.7662          & \phantom{-}0.5700          \\
                \textsc{Proden+Ens}                          & \phantom{-}0.2741                                               & \phantom{-}0.8559          & \phantom{-}0.6144          \\
                \textsc{Proden+AdvEns}                       & \phantom{-}0.2017                                               & \phantom{-}0.6435          & \phantom{-}0.4506          \\
                $\blacktriangleright$ \textsc{RobustPll+Ens} & \phantom{-}\textbf{0.5560}                                      & \phantom{-}\textbf{0.8866} & \phantom{-}\textbf{0.9996} \\
                \bottomrule
            \end{tabular}
        \end{small}
    \end{center}
\end{table}

Out-of-distribution examples (OOD) are instances that are not represented within the dataset.
Since all methods output a discrete probability distribution over known class labels, we evaluate the entropy of the predicted probability outputs.
Test-set instances should receive minimal predictive entropy, that is, the model is confident about one label, while the OOD examples should receive maximal predictive entropy, that is, no known class label matches the features.
Robust algorithms should maximize the distance between the predictive entropies on the test and OOD sets.
This is especially challenging in PLL as no exact ground truth is available.

Table~\ref{tab:ood} shows the differences in the normalized entropies (range 0 to 1) on the test and OOD samples for all methods.
All methods are trained on the MNIST train set, evaluated on the MNIST test set, and evaluated on the NotMNIST test set.
Samples from the NotMNIST test set contain letters instead of digits and are hence OOD examples.
We measure the differences between the entropies on the test and OOD set using the area between the empirical CDFs, the value of the Kolmogorov-Smirnov statistic, and the maximum-mean discrepancy with RBF kernel using the median distance heuristic to set the kernel's parameter.
We highlight the best (and close-to-best) values and consider non-ensemble and ensemble methods separately for fairness.
Positive values indicate that test predictions are taken more confidently and negative values indicate that OOD predictions are taken more confidently.

Our methods (\textsc{RobustPll} and \textsc{RobustPll+Ens}) are among the best in almost all of the three settings in Table~\ref{tab:ood}.
Some other methods even give negative values, which means that they are more sure about predicting the OOD than the test samples.
Appendix~\ref{sec:bonus-exp} also contains further results.
OOD examples mislead most of the state-of-the-art PLL methods into confidently predicting an incorrect label.
In contrast, \textsc{RobustPll+Ens} achieves almost perfect differences indicating small predictive entropies on the test set (one class label receives most of the probability mass) and high predictive entropies on the OOD set (class probabilities are almost uniformly distributed).

\subsection{Performance on Adversarial Examples}
\label{sec:adv}

\begin{table*}[tp]
    \caption{
        Average test-set accuracies ($\pm$ std.) on the real-world datasets.
        The instance features, which are min-max-normalized to the range $[0, 1]$, are corrupted using the projected gradient descent method \citep{MadryMSTV18}.
        As this attack applies to neural networks, we report only the performances of the deep learning PLL methods.
        Note that the column with $\varepsilon = 0.0$ is identical to the last column of Table~\ref{tab:accuracies}.
        The triangles indicate our proposed methods.
    }
    \label{tab:adv}
    \begin{center}
        \begin{small}
            \begin{tabular}{
                >{\raggedright\arraybackslash}m{0.19\textwidth}
                >{\centering\arraybackslash}m{0.144\textwidth}
                >{\centering\arraybackslash}m{0.144\textwidth}
                >{\centering\arraybackslash}m{0.144\textwidth}
                >{\centering\arraybackslash}m{0.144\textwidth}
                >{\centering\arraybackslash}m{0.144\textwidth}
                }
                \toprule
                \multirow[c]{2}{*}[-0.075cm]{\parbox{2cm}{Deep-learning methods}} & \multicolumn{5}{c}{Corrupted real-world datasets with adversarial parameter $\varepsilon$}                                                                                                                                             \\
                \cmidrule(ll){2-6}
                                                                                  & {\centering $\varepsilon = 0.0$}                                                           & {\centering $\varepsilon = 0.1$} & {\centering $\varepsilon = 0.2$} & {\centering $\varepsilon = 0.3$} & {\centering $\varepsilon = 0.4$} \\ 
                \midrule
                \textsc{Proden} (2020)                                            & \textbf{64.1} ($\pm$ 7.4)                                                                  & 28.9 ($\pm$ 7.1)                 & 22.6 ($\pm$ 5.7)                 & 18.8 ($\pm$ 5.2)                 & 17.2 ($\pm$ 5.9)                 \\ 
                \textsc{Proden+L2}                                                & \textbf{64.1} ($\pm$ 7.5)                                                                  & 28.4 ($\pm$ 7.6)                 & 22.5 ($\pm$ 6.1)                 & 19.2 ($\pm$ 5.3)                 & 17.2 ($\pm$ 5.3)                 \\ 
                \textsc{Proden+Edl}                                               & 49.6 ($\pm$ 21.0)                                                                          & \textbf{36.2} ($\pm$ 14.8)       & \textbf{32.0} ($\pm$ 14.0)       & \textbf{29.0} ($\pm$ 13.8)       & \textbf{27.1} ($\pm$ 13.0)       \\ 
                \textsc{Rc} (2020)                                                & 62.1 ($\pm$ 8.9)                                                                           & 29.0 ($\pm$ 6.5)                 & 21.3 ($\pm$ 6.4)                 & 17.9 ($\pm$ 6.0)                 & 14.7 ($\pm$ 5.3)                 \\ 
                \textsc{Cc} (2020)                                                & 43.8 ($\pm$ 31.2)                                                                          & 20.2 ($\pm$ 14.9)                & 14.6 ($\pm$ 11.2)                & 11.8 ($\pm$ 9.1)                 & \phantom{0}9.8 ($\pm$ 7.7)       \\ 
                \textsc{Valen} (2021)                                             & 53.8 ($\pm$ 9.0)                                                                           & 25.4 ($\pm$ 7.8)                 & 19.6 ($\pm$ 6.9)                 & 17.2 ($\pm$ 6.5)                 & 15.4 ($\pm$ 6.5)                 \\ 
                \textsc{Cavl} (2022)                                              & 61.4 ($\pm$ 6.7)                                                                           & 25.8 ($\pm$ 7.2)                 & 19.3 ($\pm$ 5.7)                 & 16.4 ($\pm$ 5.2)                 & 13.9 ($\pm$ 4.8)                 \\ 
                \textsc{Pop} (2023)                                               & \textbf{64.1} ($\pm$ 7.5)                                                                  & 28.6 ($\pm$ 7.1)                 & 22.3 ($\pm$ 6.3)                 & 18.8 ($\pm$ 5.0)                 & 16.7 ($\pm$ 4.9)                 \\ 
                \textsc{CroSel} (2024)                                            & 41.9 ($\pm$ 30.1)                                                                          & 22.6 ($\pm$ 16.8)                & 16.3 ($\pm$ 12.6)                & 13.4 ($\pm$ 10.3)                & 11.5 ($\pm$ 8.6)                 \\ 
                $\blacktriangleright$ \textsc{RobustPll}                          & 59.5 ($\pm$ 6.8)                                                                           & \textbf{40.3} ($\pm$ 12.0)       & \textbf{31.8} ($\pm$ 10.3)       & \textbf{27.4} ($\pm$ 9.3)        & 23.8 ($\pm$ 8.4)                 \\ 
                \midrule
                \textsc{Prod.+Dropout}                                            & 65.0 ($\pm$ 8.1)                                                                           & 30.7 ($\pm$ 6.1)                 & 23.4 ($\pm$ 4.8)                 & 19.8 ($\pm$ 5.0)                 & 17.9 ($\pm$ 5.4)                 \\ 
                \textsc{Prod.+Ens}                                                & 65.8 ($\pm$ 8.3)                                                                           & 42.8 ($\pm$ 6.9)                 & 33.1 ($\pm$ 8.4)                 & 27.4 ($\pm$ 8.7)                 & 24.7 ($\pm$ 10.3)                \\ 
                \textsc{Prod.+AdvEns}                                             & \textbf{66.7} ($\pm$ 9.0)                                                                  & \textbf{48.7} ($\pm$ 7.0)        & 37.1 ($\pm$ 7.6)                 & 30.5 ($\pm$ 8.7)                 & 26.6 ($\pm$ 9.9)                 \\ 
                $\blacktriangleright$ \textsc{RPll+Ens}                           & 63.6 ($\pm$ 7.8)                                                                           & \textbf{51.0} ($\pm$ 10.3)       & \textbf{42.0} ($\pm$ 10.2)       & \textbf{37.0} ($\pm$ 9.4)        & \textbf{33.3} ($\pm$ 9.1)        \\ 
                \bottomrule
            \end{tabular}
        \end{small}
    \end{center}
\end{table*}

In recent years, many attacks on neural networks have been discussed in the literature \citep{SzegedyZSBEGF13,GoodfellowSS14,MoosaviDezfooli16,MadryMSTV18}.
Using the \emph{projected gradient descent} (PGD; \citealt{MadryMSTV18}), we modify all test set examples, which are min-max-normalized to the range $[0, 1]$, by iteratively adding $\alpha := \varepsilon / 10$ times $\sgn \nabla_{x} f(x; \vc{\theta})$ to an instance's features $x \in \X$ and then projecting the newly obtained features back to an $\varepsilon$-ball around the original feature values $x$.
We repeat those steps $T = 10$ times.
The perturbed instances remain similar but moving against the gradient with respect to an instance's features decreases prediction performance rapidly.

Table~\ref{tab:adv} shows how all neural-network-based methods perform for varying values of the adversarial parameter $\varepsilon \in \lbrace 0.0, 0.1, 0.2, 0.3, 0.4 \rbrace$ on the real-world datasets.
A value of $\varepsilon = 0.0$ indicates no added adversarial noise.
The first column in Table~\ref{tab:adv} therefore matches the last column of Table~\ref{tab:accuracies}.
For values of $\varepsilon \geq 0.1$, \textsc{RobustPll} and \textsc{Proden+Edl} perform best among all non-ensemble techniques.
Among the ensemble techniques, our method (\textsc{RobustPll+Ens}) performs best.
In general, all ensembling techniques make \textsc{Proden} more robust against the corrupted features.
Note that \textsc{Proden+AdvEns} has an unfair advantage in the analysis in Table~\ref{tab:adv} as it is trained on adversarial examples, that is, it has access to the corrupted features during training.
Nevertheless, our ensemble method \textsc{RobustPll+Ens} is significantly better for $\varepsilon \geq 0.1$.
\textsc{RobustPll} and \textsc{RobustPll+Ens} consistently perform among the best for $\varepsilon \geq 0.1$.

In summary, our non-ensemble and ensemble methods consistently perform the best across almost all settings considered.
Our methods are robust against high PLL noise, out-of-distribution examples, and adversarial perturbations.

\section{Conclusions}
\label{sec:conclusions}
In this work, we presented a novel PLL method that leverages class activation values within the subjective logic framework.
We formally analyzed our method showing our update rule's optimality with respect to the mean-squared error and its reinterpretation in subjective logic.
We empirically showed that our approach yields more robust predictions than other state-of-the-art approaches in terms of prediction quality under high PLL noise, dealing with out-of-distribution examples, as well as handling instance features corrupted by adversarial noise.
To the best of our knowledge, we are the first to address these aspects in the PLL setting.


\section*{Acknowledgements}
This work was supported by the German Research Foundation (DFG) Research Training Group GRK 2153: \emph{Energy Status Data --- Informatics Methods for its Collection, Analysis and Exploitation} and by the KiKIT (The Pilot Program for Core-Informatics at the KIT) of the Helmholtz Association.

\bibliography{references}

\begin{appendices}

    \section{Proofs}
    \label{sec:proofs}
    This section collects all proofs of the propositions in the main text. The proof of Proposition~\ref{prop:var} is in Appendix~\ref{sec:proof-prop-var}, that of Proposition~\ref{prop:optimal} is in Appendix~\ref{sec:proof-optimal}, that of Proposition~\ref{prop:reinterpret} is in Appendix~\ref{sec:proof-reinterpret}, and that of Proposition~\ref{prop:bound} is in Appendix~\ref{sec:proof-bound}.

    \subsection{Proof of Proposition~\ref{prop:var}}
    \label{sec:proof-prop-var}
    Solving $\loss_{ij}^{\operatorname{err}} = \loss_{ij}^{\operatorname{var}}$ for the label weights $\lbl_{ij}$ yields
    \begin{align}
        \loss_{ij}^{\operatorname{err}} = \loss_{ij}^{\operatorname{var}}
         & \Leftrightarrow (\lbl_{ij} - \bar{p}_{ij})^2 = \frac{\bar{p}_{ij} (1 - \bar{p}_{ij})}{1 + \| \vc{\alpha}_{i} \|_1}      \\
         & \Leftrightarrow \lbl_{ij} = \bar{p}_{ij} \pm \sqrt{\frac{\bar{p}_{ij} (1 - \bar{p}_{ij})}{1 + \| \vc{\alpha}_{i} \|_1}} \\
         & \Leftrightarrow \lbl_{ij} = \bar{p}_{ij} \pm \sqrt{\loss_{ij}^{\operatorname{var}}} \text{.}
    \end{align}
    Since $\loss_{ij}^{\operatorname{err}} = (\lbl_{ij} - \bar{p}_{ij})^2$ reaches its minimum when $\lbl_{ij} = \bar{p}_{ij}$, we have shown the statement to be demonstrated.

    \subsection{Proof of Proposition~\ref{prop:optimal}}
    \label{sec:proof-optimal}
    The proof first shows that $\vc{\lbl}_{i}^{*}$ is a feasible solution for the constraints in~\eqref{eq:opt-label-weights} and then establishes that $\vc{\lbl}_{i}^{*}$ is indeed optimal using the Lagrangian multiplier method.

    \textbf{(Primal) Feasibility.}
    To prove our solution's feasibility, we need to show that $(i)$ $\| \vc{\lbl}_{i}^{*} \|_{1} = 1$ and $(ii)$ $\lbl_{ij}^{*} = 0$ for all $j \notin S_i$.
    Constraint $(i)$ holds as
    \begin{align}
        \sum_{j=1}^{k} \lbl_{ij}^{*}
         & = \sum_{j \in S_i} \Big( \bar{p}_{ij} + \frac{1}{|S_i|} (1 - \sum_{j' \in S_i} \bar{p}_{ij'} ) \Big)      \\
         & = \sum_{j \in S_i} \bar{p}_{ij} + \frac{1}{|S_i|} \sum_{j \in S_i} (1 - \sum_{j' \in S_i} \bar{p}_{ij'} ) \\
         & = \sum_{j \in S_i} \bar{p}_{ij} + 1 - \sum_{j' \in S_i} \bar{p}_{ij'} = 1 \text{.}
    \end{align}
    Constraint $(ii)$ follows directly from the definition of $\lbl_{ij}^{*}$ in Proposition~\ref{prop:optimal}.

    \textbf{Optimality.}
    Since the loss $\loss$ is differentiable, continuous, and convex in $\vc{\lbl}_i$, we can incorporate the constraints $(i)$ and $(ii)$ using the Lagrangian multiplier method as follows:
    \begin{align}
        \lagrange(\vc{\lbl}_{i}; \lambda_{i}) = \sum_{j=1}^{k} (\lbl_{ij} - \bar{p}_{ij})^2 + \lambda_{i} \Big( \sum_{j=1}^{k} \lbl_{ij} - 1 \Big) \text{,}
    \end{align}
    for instance $i \in [n]$.
    Constraint $(ii)$ directly determines the value of $\lbl_{ij}$ for all $j \notin S_i$.
    We then need to check the following Lagrange conditions:
    \begin{align}
        \frac{\partial}{\partial \lbl_{ij}} \lagrange(\vc{\lbl}_{i}; \lambda_{i}) = 0
         & \Leftrightarrow 2 (\lbl_{ij} - \bar{p}_{ij}) + \lambda_{i} = 0                                \\
         & \Leftrightarrow \lbl_{ij} = \bar{p}_{ij} - \frac{\lambda_{i}}{2} \label{eq:lagrange} \text{,}
    \end{align}
    for $i \in [n]$ and $j \in S_i$.
    For $j \notin S_i$, we have $\lbl_{ij} = 0$.
    Inserting~\eqref{eq:lagrange} into constraint $(i)$ yields
    \begin{align}
        \| \vc{\lbl}_{i} \|_{1} = 1
         & \Leftrightarrow \sum_{j=1}^{k} \lbl_{ij} = \sum_{j \in S_i} (\bar{p}_{ij} - \frac{\lambda_{i}}{2}) = \sum_{j \in S_i} \bar{p}_{ij} - \frac{|S_i| \lambda_i}{2} = 1 \\
         & \Leftrightarrow \lambda_i = \frac{2}{|S_i|} \Big( \sum_{j \in S_i} \bar{p}_{ij} - 1 \Big) \text{.} \label{eq:constraint}
    \end{align}
    Putting~\eqref{eq:constraint} back into~\eqref{eq:lagrange} gives us
    \begin{align}
        \lbl_{ij}^{*} = \begin{cases}
                            \bar{p}_{ij} + \frac{1}{|S_i|} (1 - \sum_{j' \in S_i} \bar{p}_{ij'} ) & \text{if $j \in S_i$,} \\
                            0                                                                     & \text{else,}           \\
                        \end{cases}
    \end{align}
    which is the optimal solution to~\eqref{eq:opt-label-weights}.
    Note that we do not need to show dual feasibility and complementary slackness as there are no inequality constraints.

    \subsection{Proof of Proposition~\ref{prop:reinterpret}}
    \label{sec:proof-reinterpret}
    We prove the statement by distinguishing two cases.
    (a)~If $i \in [n]$ and $j \notin S_i$, $\vc{\lbl}_{i}^{*} = \vc{\sublog{b}}_{i} + \vc{\sublog{a}}_{i} \sublog{u}_{i}$ is true as both sides are zero.
    (b)~If $i \in [n]$ and $j \in S_i$, it follows
    \begin{align}
        \lbl_{ij}^{*} & \overset{(i)}{=} \bar{p}_{ij} + \frac{1}{|S_i|} (1 - \sum_{j' \in S_i} \bar{p}_{ij'} )                                                                                                                                                                                                         \\
                      & \overset{(ii)}{=} \frac{\alpha_{ij}}{\| \vc{\alpha}_{i} \|_1} + \frac{1}{|S_i|} \Big(1 - \sum_{j' \in S_i} \frac{\alpha_{ij'}}{\| \vc{\alpha}_{i} \|_1} \Big)                                                                                                                                  \\
                      & \overset{(iii)}{=} \frac{f_j(\vc{x}_i; \vc{\theta}) + 1}{\| \vc{\alpha}_{i} \|_1} + \frac{1}{|S_i|} \Big(1 - \sum_{j' \in S_i} \frac{f_{j'}(\vc{x}_i; \vc{\theta}) + 1}{\| \vc{\alpha}_{i} \|_1} \Big)                                                                                         \\
                      & \overset{(iv)}{=} \frac{f_j(\vc{x}_i; \vc{\theta})}{\| \vc{\alpha}_{i} \|_1} + \frac{1}{\| \vc{\alpha}_{i} \|_1} + \frac{1}{|S_i|} \Big(1 - \frac{|S_i|}{\| \vc{\alpha}_{i} \|_1} - \sum_{j' \in S_i} \frac{f_{j'}(\vc{x}_i; \vc{\theta})}{\| \vc{\alpha}_{i} \|_1} \Big)                      \\
                      & \overset{(v)}{=} \underbrace{\frac{f_j(\vc{x}_i; \vc{\theta})}{\| \vc{\alpha}_{i} \|_1}}_{= \sublog{b}_{ij}} + \underbrace{\frac{1}{|S_i|}}_{= \sublog{a}_{ij}} \underbrace{\Big(1 - \sum_{j' \in S_i} \frac{f_{j'}(\vc{x}_i; \vc{\theta})}{\| \vc{\alpha}_{i} \|_1} \Big)}_{= \sublog{u}_{i}} \\
                      & \overset{(vi)}{=} \sublog{b}_{ij} + \sublog{a}_{ij} \sublog{u}_{i} \text{,}
    \end{align}
    where $(i)$ holds by Proposition~\ref{prop:optimal}, $(ii)$ by $\bar{\vc{p}}_{i} = \vc{\alpha}_{i} / \| \vc{\alpha}_{i} \|_1$, $(iii)$ by $\vc{\alpha}_{i} = f(\vc{x}_i; \vc{\theta}) + 1$, $(iv)$ by separating summands, $(v)$ by simplifying, and $(vi)$ by the definitions in Proposition~\ref{prop:reinterpret}.
    Note that we add the factor $\mathds{1}_{\lbrace j \in S_i \rbrace}$ to $\sublog{b}_{ij}$ and $\sublog{a}_{ij}$ to combine both cases, that is, (a) $j \notin S_i$ and (b) $j \in S_i$, into a single formula.

    \subsection{Proof of Proposition~\ref{prop:bound}}
    \label{sec:proof-bound}
    The proof of Proposition~\ref{prop:bound} proceeds as follows:
    \begin{align}
        0 & \leq \| \vc{\lbl}_{i}^{(t+1)} - \vc{\lbl}_{i}^{(t)} \|_{2}^{2}                                                                                                                                                                                              \\
          & \overset{(i)}{=} \sum_{j \in S_i} \Big( \bar{p}_{ij}^{(t+1)} + \frac{1}{|S_i|} \big(1 - \sum_{j' \in S_i} \bar{p}_{ij'}^{(t+1)} \big) - \bar{p}_{ij}^{(t)} - \frac{1}{|S_i|} \big(1 - \sum_{j' \in S_i} \bar{p}_{ij'}^{(t)} \big) \Big)^2                   \\
          & \overset{(ii)}{=} \sum_{j \in S_i} \Big( \big(\bar{p}_{ij}^{(t+1)} - \bar{p}_{ij}^{(t)} \big) + \frac{1}{|S_i|} \sum_{j' \in S_i} \big( \bar{p}_{ij'}^{(t)} - \bar{p}_{ij'}^{(t+1)} \big) \Big)^2                                                           \\
          & \overset{(iii)}{=} \sum_{j \in S_i} \big(\bar{p}_{ij}^{(t+1)} - \bar{p}_{ij}^{(t)} \big)^2 + \frac{2}{|S_i|} \sum_{j \in S_i} \big(\bar{p}_{ij}^{(t+1)} - \bar{p}_{ij}^{(t)} \big) \sum_{j' \in S_i} \big(\bar{p}_{ij'}^{(t)} - \bar{p}_{ij'}^{(t+1)} \big) \\
          & \qquad + \frac{1}{|S_i|^2} \sum_{j \in S_i} \Big( \sum_{j' \in S_i} \big(\bar{p}_{ij'}^{(t)} - \bar{p}_{ij'}^{(t+1)} \big) \Big)^2                                                                                                                          \\
          & \overset{(iv)}{=} \sum_{j \in S_i} \big(\bar{p}_{ij}^{(t+1)} - \bar{p}_{ij}^{(t)} \big)^2
        - \frac{2}{|S_i|} \Big( \sum_{j \in S_i} \big(\bar{p}_{ij}^{(t+1)} - \bar{p}_{ij}^{(t)} \big) \Big)^2 + \frac{1}{|S_i|} \Big( \sum_{j' \in S_i} \big(\bar{p}_{ij'}^{(t)} - \bar{p}_{ij'}^{(t+1)} \big) \Big)^2                                                  \\
          & \overset{(v)}{=} \sum_{j \in S_i} \big(\bar{p}_{ij}^{(t+1)} - \bar{p}_{ij}^{(t)} \big)^2
        - \frac{1}{|S_i|} \Big( \sum_{j \in S_i} \big(\bar{p}_{ij}^{(t+1)} - \bar{p}_{ij}^{(t)} \big) \Big)^2                                                                                                                                                           \\
          & \overset{(vi)}{\leq} \sum_{j \in S_i} \big(\bar{p}_{ij}^{(t+1)} - \bar{p}_{ij}^{(t)} \big)^2                                                                                                                                                                \\
          & \overset{(vii)}{\leq} \sum_{j=1}^{k} \big(\bar{p}_{ij}^{(t+1)} - \bar{p}_{ij}^{(t)} \big)^2                                                                                                                                                                 \\
          & = \| \bar{\vc{p}}_{i}^{(t+1)} - \bar{\vc{p}}_{i}^{(t)} \|_{2}^{2} \text{,}
    \end{align}
    where $(i)$ holds by Proposition~\ref{prop:optimal}, $(ii)$ by reordering, $(iii)$ by the binomial theorem, $(iv)$ by
    \begin{align}
        \sum_{j' \in S_i} \big(\bar{p}_{ij'}^{(t)} - \bar{p}_{ij'}^{(t+1)} \big) = - \sum_{j' \in S_i} \big(\bar{p}_{ij'}^{(t+1)} - \bar{p}_{ij'}^{(t)} \big) \text{,}
    \end{align}
    and $(v)$ by
    \begin{align}
        \Big( \sum_{j \in S_i} \big(\bar{p}_{ij}^{(t)} - \bar{p}_{ij}^{(t+1)} \big) \Big)^2
        = \Big( -\sum_{j \in S_i} \big(\bar{p}_{ij}^{(t+1)} - \bar{p}_{ij}^{(t)} \big) \Big)^2
        = \Big( \sum_{j \in S_i} \big(\bar{p}_{ij}^{(t+1)} - \bar{p}_{ij}^{(t)} \big) \Big)^2 \text{.}
    \end{align}
    $(vi)$ holds as $|S_i| \geq 1$ and $\sum_{j=1}^{k} \big(\bar{p}_{ij}^{(t+1)} - \bar{p}_{ij}^{(t)} \big)^2 \geq 0$ and $(vii)$ holds by including further non-negative summands within the summation.

    \subsection{Proof of Proposition~\ref{prop:optimal-ce}}
    The proof first shows that $\lbl_{ij}^{*}$ is a feasible solution for the constraints in~\eqref{eq:opt-label-weights} and then establishes that $\lbl_{ij}^{*}$ is optimal.

    \textbf{Feasibility.}
    As $\lbl_{ij}^*$ is one for exactly one class label $j \in S_i$, it holds that $\| \vc{\lbl}_i \|_1 = 1$ for $i \in [n]$.
    Also, $\lbl_{ij}^* = 0$ for $j \notin S_i$ as only class labels $j \in S_i$ can be different from zero.

    \textbf{Optimality.}
    As the cross-entropy loss $\loss_{\operatorname{CE}}$ in Section~\ref{sec:cross-entropy} is a linear combination of $\lbl_{ij}$ with coefficients $\Psi_{ij} = \psi(\| \vc{\alpha}_{i} \|_1) - \psi(\alpha_{ij})$ for fixed $i \in [n]$ and $\lbl_{ij} \in [0, 1]$, we minimize $\loss_{\operatorname{CE}}$ by assigning all label weight to the minimal coefficient $\Psi_{ij}$, that is, $\arg\min_{j \in S_i} \Psi_{ij}$.

    \section{Experiments}
    \label{sec:bonus-exp}
    This section augments Section~\ref{sec:exp} presenting more details of our experimental setup and results.
    This includes the hyperparameter values of all methods as well as more results on how all methods handle adversarially corrupted instances.

    \begin{figure*}[t]
        \centering
        \includegraphics[width=\textwidth]{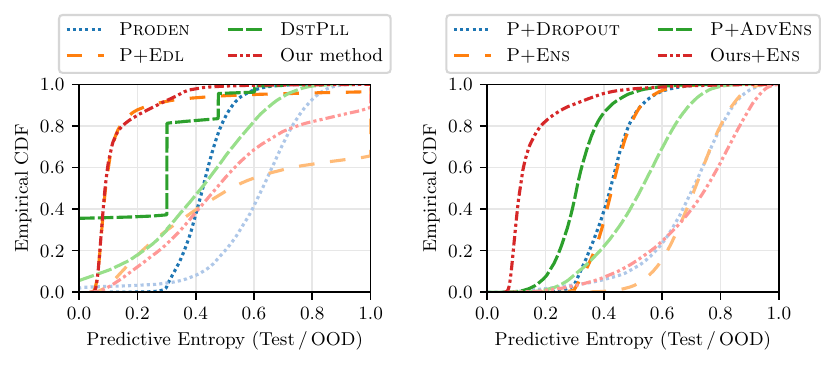}
        \caption{
            Empirical CDF of the normalized entropy (range 0 to 1) of predictions on MNIST (darker color) and NotMNIST (lighter color) for models trained on MNIST.
            The left plot shows the four best non-ensemble approaches according to Table~\ref{tab:ood} (highest metrics).
            We exclude methods that are too similar, for example, \textsc{Proden-L2} and \textsc{Rc} behave similarly to \textsc{Proden}, which is shown.
            All methods' performances can be observed in Table~\ref{tab:ood}.
            The right plot shows the predictive entropy of all four ensemble approaches.
            Our ensemble approach is most certain about predictions on the test set (top-left corner) while being one of the approaches that is the most uncertain about out-of-distribution examples (bottom-right corner).
        }
        \label{fig:ood}
    \end{figure*}

    \subsection{Hyperparameters}
    This section lists all methods, which are benchmarked in the main text, together with their respective hyperparameter choices.
    We set all hyperparameters as recommended by the respective authors.
    There are 14 non-ensemble and four ensemble methods.
    The non-ensemble methods and their hyperparameters are:
    \begin{itemize}
        \item \textsc{PlKnn} \citep{HullermeierB06}: We use $k = 10$ nearest neighbors.
        \item \textsc{PlSvm} \citep{NguyenC08}: We use the \textsc{Pegasos} optimizer \citep{Shalev-ShwartzSS07} and $\lambda = 1$.
        \item \textsc{Ipal} \citep{ZhangY15a}: We use $k = 10$ neighbors, $\alpha = 0.95$, and 100 iterations.
        \item \textsc{PlEcoc} \citep{ZhangYT17}: We use $L = \lceil 10 \log_2(l) \rceil$ and $\tau = 0.1$.
        \item \textsc{Proden} \citep{LvXF0GS20}: For a fair comparison, we use the same base model for all neural-network-based approaches.
              We use a standard $d$-300-300-300-$l$ MLP \citep{werbos1974beyond} with ReLU activations, batch normalizations, and softmax output.
              We choose the \emph{Adam} optimizer for training and train for a total of 200 epochs.
        \item \textsc{Proden+L2} \citep{HoerlK00,LvXF0GS20}: We use the same base model and settings as \textsc{Proden} with additional L2 weight regularization.
        \item \textsc{Proden+Edl} \citep{SensoyKK18,LvXF0GS20}: We use the \textsc{Proden} model to disambiguate the candidate labels with the same settings as above.
              Then, we use the evidential-learning algorithm by~\citet{SensoyKK18} in a supervised manner.
        \item \textsc{Rc} \citep{FengL0X0G0S20}: We use the same base model and settings as \textsc{Proden}.
        \item \textsc{Cc} \citep{FengL0X0G0S20}: We use the same base model and settings as \textsc{Proden}.
        \item \textsc{Valen} \citep{XuQGZ21}: We use the same base model and settings as \textsc{Proden}.
        \item \textsc{Cavl} \citep{ZhangF0L0QS22}: We use the same base model and settings as \textsc{Proden}.
        \item \textsc{Pop} \citep{0009LLQG23}: We use the same base model and settings as \textsc{Proden}.
              Also, we set $e_0 = 0.001$, $e_{end} = 0.04$, and $e_s = 0.001$.
        \item \textsc{CroSel} \citep{crosel2024}: We use the same base model and settings as \textsc{Proden}.
              We use 10 warm-up epochs using \textsc{Cc} and $\lambda_{cr} = 2$.
              We abstain from using the data augmentations discussed in the paper for a fair comparison of the base approach.
        \item \textsc{DstPll} \citep{fuchs2024partiallabel}: We use $k = 20$ neighbors and a variational auto-encoder to reduce the feature dimensionality as recommended by the authors.
        \item \textsc{RobustPll} (our method): We use the same base model and settings as \textsc{Proden}.
              The parameter $\lambda_t$ is set to $\min(2t/T, 1)$ with $T=200$ epochs.
    \end{itemize}
    The four ensemble methods and their hyperparameters are:
    \begin{itemize}
        \item \textsc{Proden+Dropout} \citep{SrivastavaHKSS14,LvXF0GS20}: We use the \textsc{Proden} model with additional Monte-Carlo dropout. The dropout layer is also active during inference. We repeat the predictions 1000 times to estimate the uncertainty involved across all dropout networks.
        \item \textsc{Proden+Ens} \citep{Lakshminarayanan17,LvXF0GS20}: We use an ensemble of 5 \textsc{Proden} models.
        \item \textsc{Proden+AdvEns} \citep{LvXF0GS20,Lakshminarayanan17}: We use an ensemble of 5 \textsc{Proden} models that are trained on adversarially corrupted instance features.
        \item \textsc{RobustPll+Ens} (our method): We use an ensemble of 5 \textsc{RobustPll} models.
    \end{itemize}

    \subsection{Adversarial Perturbations}

    To complement Table~\ref{tab:ood} in the main text, Figure~\ref{fig:ood} provides the empirical cumulative distribution functions on the test and OOD set of the four best non-ensemble methods (regarding Table~\ref{tab:ood}) on the left and of the four ensemble approaches on the right.
    The empirical CDFs of the entropies are normalized to a range between zero and one.
    The dark-colored lines represent the entropy CDFs of the predictions on the MNIST test set.
    The light-colored lines represent the entropy CDFs of the predictions on the NotMNIST test set (OOD).
    The left plot shows the four best non-ensemble approaches according to Table~\ref{tab:ood} (highest metrics).
    We exclude methods that are too similar, for example, \textsc{Proden-L2} and \textsc{Rc} behave similarly to \textsc{Proden}, which is shown.
    All methods' performances can be observed in Table~\ref{tab:ood}.
    The right plot shows the predictive entropy of all four ensemble approaches.
    Our ensemble approach is most certain about predictions on the test set (top-left corner) while being one of the approaches that is the most uncertain about out-of-distribution examples (bottom-right corner).

\end{appendices}

\end{document}